\documentclass[conference]{IEEEtran}
\IEEEoverridecommandlockouts
\usepackage[bookmarks=false]{hyperref}
\usepackage{cite}
\usepackage{amsmath,amssymb,amsfonts}
\usepackage{amsthm}
\usepackage{algorithm}
\usepackage{algpseudocode}
\usepackage{graphicx}
\usepackage{textcomp}
\usepackage{subcaption}
\usepackage{multirow}
\usepackage{xcolor}
\usepackage{mathtools}
\usepackage{hyperref}

\theoremstyle{definition}
\newtheorem{definition}{Definition}

\def\BibTeX{{\rm B\kern-.05em{\sc i\kern-.025em b}\kern-.08em
    T\kern-.1667em\lower.7ex\hbox{E}\kern-.125emX}}

\DeclarePairedDelimiter\abs{\lvert}{\rvert}

\begin{document}

\title{Fair Meta-Learning For Few-Shot Classification
}


\author{\IEEEauthorblockN{Chen Zhao, Changbin Li, Jincheng Li, Feng Chen}
\IEEEauthorblockA{\textit{Department of Computer Science} \\
\textit{The University of Texas at Dallas}\\
Richardson Texas, USA \\
\{chen.zhao, changbin.li, jincheng.li, feng.chen\}@utdallas.edu}

}

\maketitle

\begin{abstract}
Artificial intelligence nowadays plays an increasingly prominent role in our life since decisions that were once made by humans are now delegated to automated systems. A machine learning algorithm trained based on biased data, however, tends to make unfair predictions. Developing classification algorithms that are fair with respect to protected attributes of the data thus becomes an important problem. Motivated by concerns surrounding the fairness effects of sharing and few-shot machine learning tools, such as the Model Agnostic Meta-Learning \cite{Finn-ICML-2017-(MAML)} framework, we propose a novel fair fast-adapted few-shot meta-learning approach that efficiently mitigates biases during meta-train by ensuring controlling the decision boundary covariance that between the protected variable and the signed distance from the feature vectors to the decision boundary. Through extensive experiments on two real-world image benchmarks over three state-of-the-art meta-learning algorithms, we empirically demonstrate that our proposed approach efficiently mitigates biases on model output and generalizes both accuracy and fairness to unseen tasks with a limited amount of training samples. 
\end{abstract}

\begin{IEEEkeywords}
decision boundary covariance, statistical parity ,few-shot, meta-learning
\end{IEEEkeywords}

\section{Introduction}
\begin{figure}
    \centering
    \includegraphics[width = \linewidth]{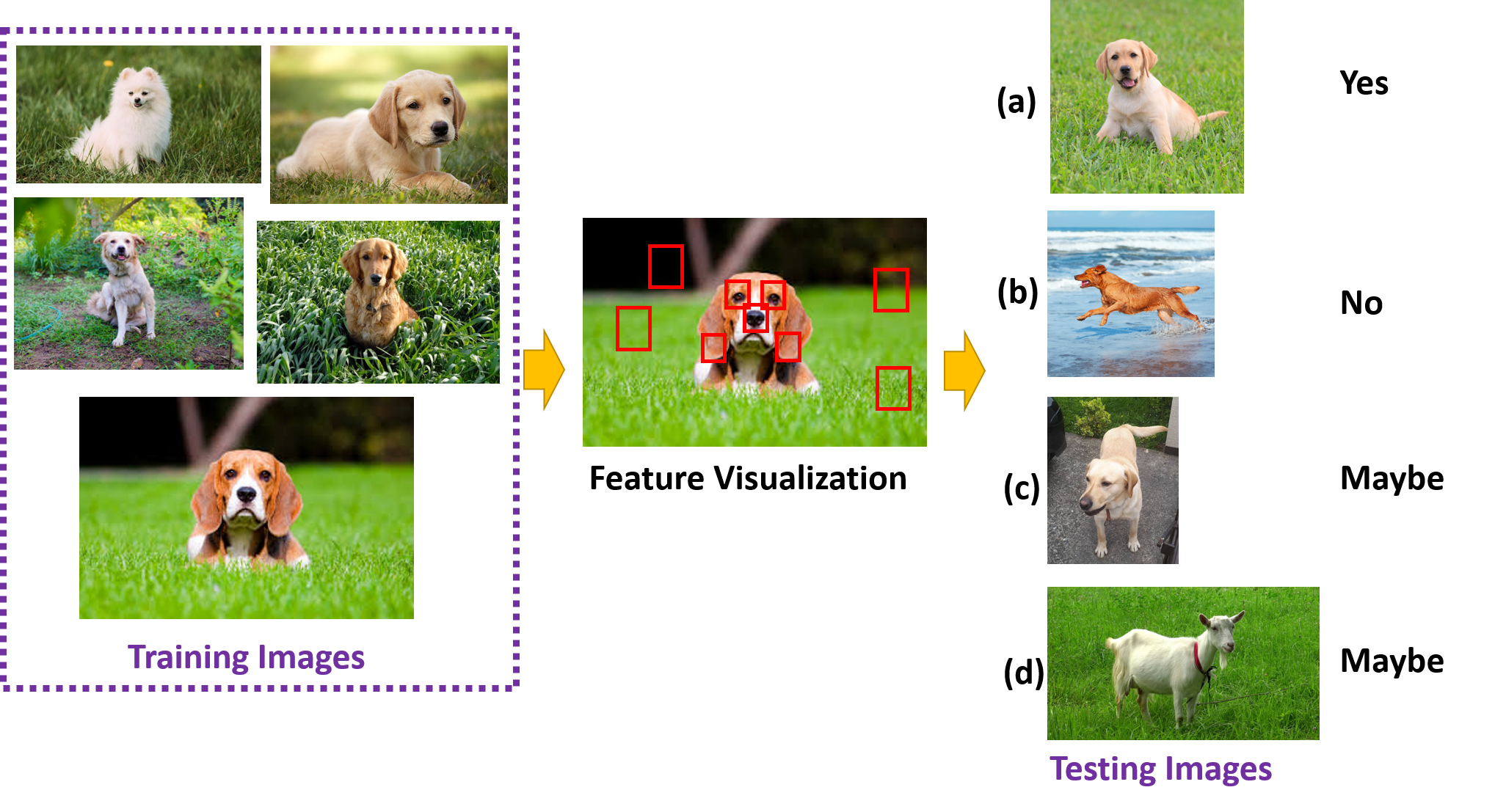}
    \caption{An example of biased image classification problem. A dog classifier is trained using images of dogs lying on the grass. Features learned from the neural network, however, are not fully concentrated on the target objects (\textit{i.e.} dogs). As a consequence, the biased learner leads to unfair or uncertain decision-makings: (a) a dog on the grass; (b) a dog running on the beach; (c) a dog on the stone step; (d) a goat on the grass. }
    \label{fig:dogs}
\end{figure}

In data mining and machine learning, the information system is becoming increasingly reliant on statistical inference and learning to give automated prediction and decision-making to solve regression and classification problems. Biased historical data or data containing biases, however, are often learned and thus lead to results with undesirability, inaccuracy, and even illegality. In recent years, there are increasing numbers of news reported that human bias is revealed in an artificial intelligence system applied by high-tech companies. \cite{Barr-Google-Gorillas-2015} reported that a picture of two African Americans was automatically tagged as ``Gorillas'' by \textit{Google Photos}. A 2016 study \cite{Ingold-Amazon-2016} found that the data-driven system developed by \textit{Amazon} that used to determine the neighborhoods in which to offer free same-day delivery is highly biased and unfair to African American communities due to the stark disparities in the demographic makeup of neighborhoods: white residents were more than twice as likely as African American residents to live in one of the qualifying neighborhoods. Critics have voiced that human bias potentially has an influence on nowadays technology, which leads to outcomes with unfairness. Another example for biased image classification problem is shown in Figure \ref{fig:dogs}: a dog classifier is trained with images of dogs lying on the grass. The training process goes through a feature extractor, but the captured features used for classifier training are not totally concentrated on target objects (\textit{i.e.} dogs). As a consequence, the decision-making accuracy for testing images does not turn out well. To investigate the reason, we deduce that there is a non-negligible relationship between the predicted outcome and the protected feature (\textit{i.e.} grass in this example), which leads to an unfair result. 

To ameliorate this unfairness problem, one may attempt to make the automated decision-maker blind to the protected attributes \cite{Ren-NeurIPS-2019}. This however, is difficult, as many attributes may be correlated with the protected one \cite{Zemel-ICML-2013}. With biased input, the main goal of training an unbiased model is to make the output fair. In other words, the predicted outcomes are statistically independent on protected variables. Statistical parity, also known as group fairness, ensures that the overall proportion of members in a protected group receiving predictions (\textit{i.e.} positive/negative classification) are identical to the proportion of the population as a whole.

\begin{figure*}[!htbp]
    \centering
    \includegraphics[width=\textwidth]{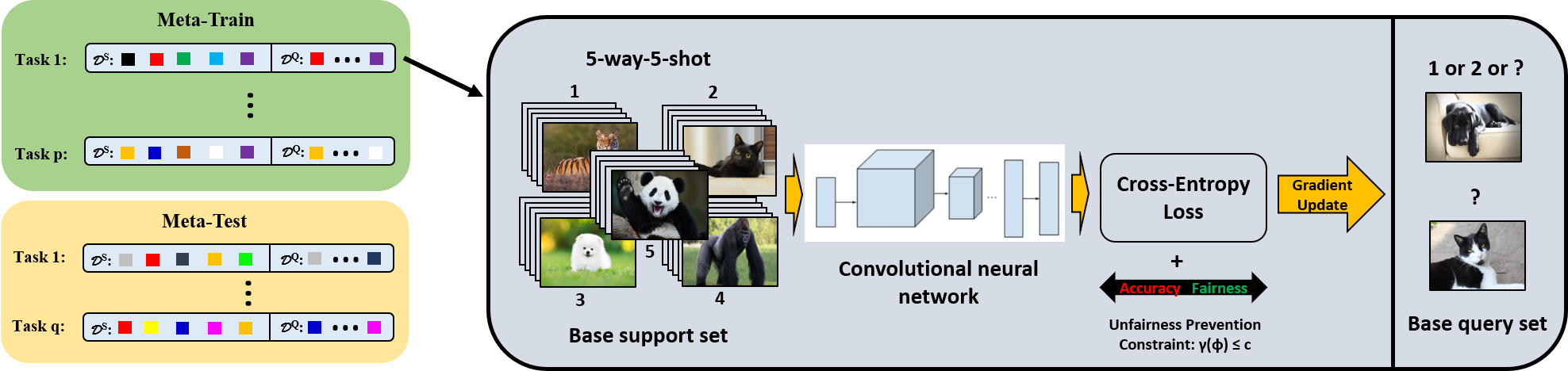}
    \caption{An overview of our proposed bias controlling approach in few-shot meta-learning. (Left) The meta-learning framework include two processes: meta-train and meta-test, where each includes multiple tasks. (Right) Taking 5-way-5-shot classification problem as an example, $\gamma(\phi)$ refers to a discrimination measure function, and $c$ is a predefined small threshold to account for a degree of randomness in the decision making process and sampling. The constraint here is considered to ensure no discrimination on the prediction model for each task.}
    \label{fig:overvoew}
\end{figure*}

To the best of our knowledge, unfortunately, the majority of existing fairness-aware machine learning algorithms are under the assumption of giving abundant training examples. Learning quickly, however, is another significant hallmark of human intelligence. In meta-learning, also known as learning to learn, the goal of trained model is to quickly learn a new task from a small amount of new data (\textit{i.e.} few-shot), and the model is trained by the meta-leaner to be able to learn on a large number of different tasks \cite{Finn-ICML-2017-(MAML)}. In contrast to traditional machine learning algorithms, such as multi-task learning \cite{Liu-UAI-2009,Ruder-arXiv-2017,Zhang-NSR-2018, Zhao-ICDM-2019} and transfer learning \cite{Dong-ECMLPKDD-2018,Hsu-ICLR-2018,Ganin-ICML-2015,Pan-TKDE-2010}, meta-learning framework has advantages: (1) it learns across tasks where each task takes one or few samples as input; (2) it therefore efficiently speeds up model adaptation (3) and generalizes accuracy to unseen tasks. 

The overall idea of existing methods of meta-learning, however, is to train a model which is capability of generalizing accuracy, rather than fairness, to unseen data tasks. But techniques for unfairness prevention and bias control in the few-shot meta-learning study are challenging and rarely touched. To ensure prediction without biases, the main contribution to this paper is that we feed each support set of a task with unified group fairness constraints and minimize meta-loss overall episodes. Specifically, we mitigate biases in each episode during meta-training by controlling the decision boundary covariance \cite{Zafar-AISTATS-2017} which is defined as the covariance between the protected attribute and the signed distance from the feature vectors to the decision boundary.  A value of zero signifying no dependency or attribute effect. Our experimental results demonstrate our approach is capability of controlling bias and decreasing loss as well as generalizing both to unseen tasks. In the context of classification, for example, as shown in Figure \ref{fig:overvoew}, each support set of a task used for training contains images sampled from 5 different classes ($N=5$ ways) and each class includes 5 images ($K=5$ shots). By giving an unified meta-initialization for each task, a task specific local model parameter is learned through one or few steps gradient update of the loss function that is constrained by fairness condition. To update the meta-parameter, the generalization error, \textit{i.e.} the summation of the query loss across all tasks, is minimized. In summary, the main contributions of this paper are listed:

\begin{itemize}
    \item For the first time the issue of bias control in meta-learning multi-class classification problem is applied to image data sets. We mitigate biases by controlling the decision boundary covariance.
    \item We develop a novel algorithm to solve this constrained classification problem under the Model Agnostic Meta-Learning (MAML) few-shot framework.
    \item We validate the performance of our proposed approach of controlling biases on three state-of-the-art meta-learning techniques through extensive experiments based on real-world data sets. Our results demonstrate the proposed approach is capability of mitigating biases and generalizing both accuracy and fairness to unseen tasks, with the input training data is minimal.
\end{itemize}

In Section 2, some related works are referred. In Section 3, we see how unfairness is important in a machine learning model by introducing a simple causal based knowledge graph and how a statistical parity constraint, i.e. decision boundary covariance, is able to be used for bias-control in a single task. In Section 4, the fair few-shot meta-learning problem is formulated and how to solve it by applying the Model-Agnostics Meta-Learning framework is presented in detail. In Section 5, to validate the proposed approach, we conduct experiments by using two real-world benchmarks and three cutting-edge techniques, and we conclude this paper in Section 6.


\section{Related Work}
In recent years, researches involving processing biased data became increasingly significant. Fairness-aware in data mining is classified into unfairness discovery and prevention. Based on the taxonomy by tasks, it can be further categorized to classification \cite{Feldman-KDD-2015, Zafar-AISTATS-2017, Fish-arxiv-2015, Hardt-NIPS-2016,Kamishima-DMKD-w-2018,Hajian-DMKD-2015, Agarwal-ICML-2018, Oneto-AIES-2019, wang2019}, regression \cite{Calders-ICDM-2013,Zhao-ICDM-2019,Berk-FATML-2018,Komiyama-ICML-2018,PerezSuay-ECMLPKDD-2017}, clustering \cite{Gondek-ICDM-2004,Gondek-KDD-2005}, recommendation \cite{Kamishima-ICDM-w-2016,Kamishima-RR-w-2017,Singh-KDD-2018} and dimension reduction \cite{Bolukbasi-NIPS-2016}. 

\subsection{Unfairness Prevention in Classification}
Majority of works in unfairness prevention is concentrated on data classification. According to approaches studied in fairness, bias-prevention in classification is subcategorized into pre-processing\cite{Feldman-KDD-2015}, in-processing \cite{Zafar-AISTATS-2017, Fish-arxiv-2015} and post-processing \cite{Hardt-NIPS-2016,Kamishima-DMKD-w-2018,Hajian-DMKD-2015}. Recent works \cite{Agarwal-ICML-2018} and \cite{Oneto-AIES-2019} developed new approaches resulting in increasing the binary classification accuracy through reduction of fair classification to a sequence of cost-sensitive problems and applications of multi-task techniques with convex fairness constraints, respectively. 

Non-discrimination (unfairness-free) can be defined as follows: (1) people that are similar in terms of non-sensitive characteristics should receive similar predictions, and (2) differences in predictions across groups of people can only be as large as justified by non-sensitive characteristics\cite{Zliobaite-arXiv-2015}. The first condition is related to direct discrimination. The second condition ensures that there is no indirect discrimination, also referred to as redlining. These types of discrimination (direct and indirect) are supported by two legal frameworks applied in large bodies of cases, disparate treatment and disparate impact\cite{Barocas-CLR-2016}. The disparate treatment framework enforces procedural fairness, namely, the equality of treatments that prohibits the use of the protected attribute in the decision process. The disparate impact framework guarantees outcome fairness, namely, the equality of outcomes among protected groups \cite{Zhang-AAAI-2018}. Many of the prior studies, however, suffer from one or more of the following limitations: (i) they are restricted to a narrow range of classifiers, (ii) they only accommodate a single, binary sensitive attribute, and (iii) they cannot eliminate disparate treatment and disparate impact simultaneously. To overcome such limitations, in this paper, we consider the measure of decision boundary fairness \cite{Zafar-AISTATS-2017}, which enables us to ensure fairness with respect to one or more sensitive attributes, in terms of both disparate treatment and disparate impact.

\subsection{Few-shot Meta-learning}
To the best of our knowledge, the majority of existing fairness-aware machine learning algorithms are under the assumption of giving abundant training examples. Learning quickly, however, is another significant hallmark of human intelligence. Much efforts have been devoted to overcome the data efficiency issue. One popular category of few-shot learning techniques is distance metric learning based method, which addresses the few-shot classification problem by ``learning to compare”. The intuition is that if a model can determine the similarity of two images, it can classify an unseen input image with the labeled instances. \cite{Vinyals-NIPS-2016-(MatchingNet)} introduced Matching Networks which employed ideas from k-nearest neighbors algorithm and metric learning based on a bidirectional Long-Short Term Memory (LSTM) to encode in the context of the support set. Prototypical networks \cite{Snell-NIPS-2017-(ProtoNet)} learn a metric space in which classification is able to be performed by computing Euclidean distances to prototype representations of each class. In addition, gradient descent based algorithms, such as  \cite{Finn-ICML-2017-(MAML),Ravi-ICLR-2017,Finn-NIPS-2018,Nichol-arXiv-Reptile-2018,Rusu-ICLR-2019,Antoniou-ICLR-2019}, aim to learn good model initialization so that the meta-loss is minimum. 

The overall idea of these state-of-the-art is to train a meta-learning model which is capability of generalizing accuracy, but less attention on fairness generalization to unseen data tasks. In this paper, our proposed approach makes up for this regret of unfairness prevention using few-shot meta-learning techniques in multi-class classification problems.

\begin{figure}
    \centering
    \includegraphics[width=0.5\linewidth]{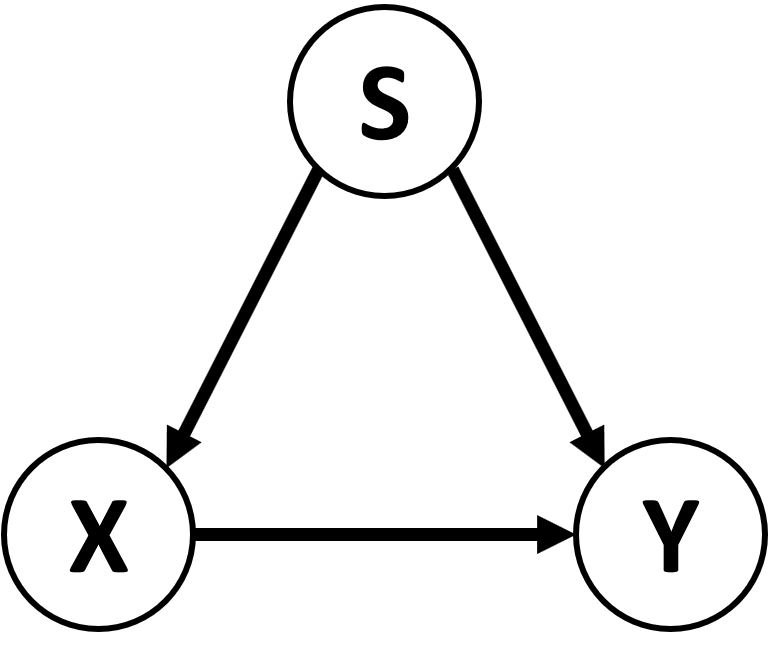}
    \caption{A simple diagram demonstrates the causal relationship in fairness learning. $X,S,$ and $Y$ represent input feature, the protected attribute and target outcome, respectively.}
    \label{fig:simpleCausal}
\end{figure}

\section{Modeling of Fairness Based on Causal Knowledge Graph}
In this section, we first present how unfairness/bias affects decision-making by introducing a simple causal knowledge graph and then explain the mechanism of mitigating bias in a single task using the decision boundary covariance. 

\subsection{Causation in Fairness Learning}
To understand how past decisions may bias a prediction model, we must first understand how the protected attribute may have affected the outcome by answering such questions:  What would this outcome have been under different protected values? How would the outcome change if the protected attribute were changed, all else being equal? These questions are core to the mission of learning fair systems which aim to inform decision-making. 

Unfairness can be broadly partitioned into two types: direct and indirect. The directed bias is concerned with settings where individuals received less favorable treatments on the basis of the protected attribute. The indirect one is concerned with individuals who receive treatments on the basis of inadequately justified factors that are somewhat related with the protected attribute \cite{Zhang-AAAI-2018}.

For simplicity, we consider one binary protected attribute (\textit{e.g.} white and black) in this work. However, our ideas can be easily extended to many protected attributes with multiple levels. Let $\mathcal{Z=X \times S\times Y}$ be the data space, where $\mathcal{X}\subset \mathbb{R}^n$ is an input space, $\mathcal{S} = \{0,1\}$ is a protected space, and $\mathcal{Y} = \{1,2,...,N\}$ is an output space for multi-class classification where $N$ is the number of classes. We consider a single task data $\mathcal{D} = \{(\mathbf{x}_i,y_i,s_i)\}_{i=1}^{h}, i=1,...h$, where $\mathbf{x}_i\in\mathbb{R}^n$ denotes the $i$-th observation, $y_i$ denotes the corresponding output, $s_i\in\{s_+, s_-\}$ represents the binary protected attribute, and $h$ is the number of observations in each task. A practical definition of fair causality is:

\begin{definition}[Fair Causality and Causal Effect]
$X$ causes $Y$ if and only if changing $X$ leads to a change in $Y$, while keeping everything else (\textit{i.e.} $S$) constant. Causal effect is defined as the magnitude by which $Y$ is changed by a unit change in $X$.
\end{definition}

Therefore, a fair prediction, shown in Figure \ref{fig:simpleCausal}, indicates there is no either direct ($S\rightarrow Y$) or indirect ($S\rightarrow X \rightarrow Y$) dependency effect of outcome on the protected attribute. These types of discrimination (direct and indirect) are supported by two legal frameworks applied in large bodies of cases throughout the disparate treatment and disparate impact \cite{Barocas-CLR-2016}. The disparate treatment framework enforces procedural fairness, namely, the equality of treatments that prohibits the use of the protected attribute in the decision process. The disparate impact framework guarantees outcome fairness, namely, the equality of outcomes among protected groups \cite{Zhang-AAAI-2018}.

To comply with disparate treatment criterion we specify that sensitive attributes are not used in decision making, \textit{i.e.} $\{\mathbf{x}_i\}_{i=1}^h$ and $\{s_i\}_{i=1}^h$ consist of disjoint feature sets. As discussed in \cite{Zafar-AISTATS-2017}, our definition of disparate impact leverages the $80\%$-rule \cite{Biddle-Gower-2005}. A decision boundary satisfies the $80\%$-rule if the ratio between the percentage of users with a particular protected attribute value having $d_\alpha(\mathbf{x})\geq 0$, where $\mathbf{\alpha}$ is the decision boundary parameter, and the percentage of users without that value having $d_\alpha(\mathbf{x})\geq 0$ is no less than 0.8 \cite{Zafar-AISTATS-2017}.

\begin{align}
\label{80-rule}
    \min \Big(\frac{P(d_\alpha(\mathbf{x})\geq 0|s=1)}{P(d_\alpha(\mathbf{x})\geq 0|s=0)}, \frac{P(d_\alpha(\mathbf{x})\geq 0|s=0)}{P(d_\alpha(\mathbf{x})\geq 0|s=1)}\Big) \geq 0.8
\end{align}

\subsection{Decision Boundary Covariance in Statistical Parity}

In this section, we introduce a measure of decision boundary fairness, which enables us to ensure fairness with respect to one or more protected attributes, in terms of both disparate treatment and disparate impact. The decision boundary covariance (DBC) which measures the decision boundary (un)fairness is defined as 

\begin{definition}[Decision Boundary Covariance \cite{Zafar-AISTATS-2017}]
The covariance between the protected variables $\mathbf{s}=\{s_i\}_{i=1}^h$ and the signed distance from the feature vectors to the decision boundary, $d_\mathbf{\alpha}(\mathbf{x}) = \{d_\mathbf{\alpha}(\mathbf{x}_i)\}_{i=1}^h$, where $\mathbf{\alpha}$ is the decision boundary parameter.
\begin{align}
\label{dbc definition}
    DBC(\mathbf{s}, d_\mathbf{\alpha}(\mathbf{x})) &= \mathbb{E}[(\mathbf{s-\Bar{s}})d_\mathbf{\alpha}(\mathbf{x})] - \mathbb{E}[\mathbf{s-\Bar{s}}]\Bar{d}_\mathbf{\alpha}(\mathbf{x}) \nonumber\\
    &\approx \frac{1}{h}\sum_{i=1}^h (\mathbf{s}_i-\mathbf{\Bar{s}})d_\mathbf{\alpha}(\mathbf{x})
\end{align}
\end{definition}

where $\mathbb{E}[\mathbf{s-\Bar{s}}]\Bar{d}_\mathbf{\alpha}(\mathbf{x})$ is cancels out since $\mathbb{E}[\mathbf{s-\Bar{s}}]=0$. Taking linear model as an example, the decision boundary is simply the hyperplane defined by $\mathbf{\alpha}^T\mathbf{x}=0$. Then the DBC reduces to $\frac{1}{h}\sum_{i=1}^h (\mathbf{s}_i-\mathbf{\Bar{s}})\mathbf{\alpha}^T\mathbf{x}$.

\begin{figure}
    \centering
    \includegraphics[width = \linewidth]{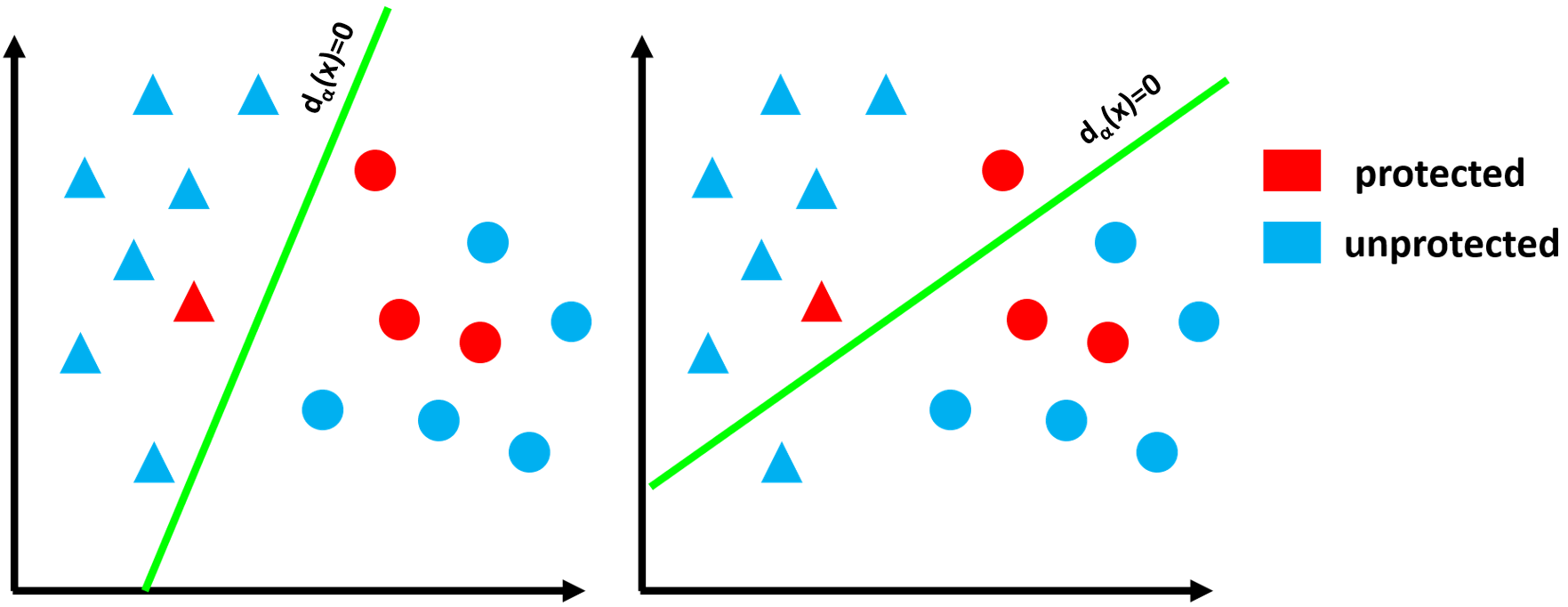}
    \caption{An example of fair binary classification (unfair (left) and fair (right) classifier) by controlling the decision boundary covariance between the protected variable and the signed distance from from user's feature to the decision boundary.  }
    \label{fig:dbc}
\end{figure}

An example of fair binary classification with a linear decision boundary is given in Figure \ref{fig:dbc}. Red markers represent the protected group (\textit{i.e.} $s=1$) and blue ones are the unprotected group (\textit{i.e.} $s=0$). In the left of Figure \ref{fig:dbc}, we calculate $P(d_\alpha(\mathbf{x})\geq 0|s=1) = 1/4 = 0.25$, where $d_\alpha(\mathbf{x})\geq 0$ indicates the triangle class and $P(d_\alpha(\mathbf{x})\geq 0|s=0) = 7/(16-4) = 0.583$. By applying the $80\%$-rule indicated in Eq.(\ref{80-rule}), the disparate impact value of the left classifier is $0.25/0.583 = 0.43$, which is lower than the threshold of $0.8$ and returns an unfair classification prediction. Similarly, in the right case of Figure \ref{fig:dbc}, however,  $P(d_\alpha(\mathbf{x})\geq 0|s=1) = P(d_\alpha(\mathbf{x})\geq 0|s=0) = 0.5$ and thus the disparate impact is $1.0$. Note that, if a decision boundary satisfies the $100\%$-rule, \textit{i.e.}
\begin{align*}
    P(d_\alpha(\mathbf{x})\geq 0|s=1) = P(d_\alpha(\mathbf{x})\geq 0|s=0)
\end{align*}
then the empirical covariance will be approximately zero for a sufficiently large training set.

\section{Fair Meta-Learning}
Meta-learning for few-shot learning aims to train a meta-learner which is able to learn on a large number of various tasks from a small amount of data. MAML (Model-Agnostic Meta-Learning) proposed by \cite{Finn-ICML-2017-(MAML)} is one of the popular gradient based meta-learning frameworks, which leads to state-of-the-art performance and fast adaptation to unseen tasks. To generalize fairness in a classification problem with minimal samples, we propose a novel approach by modifying MAML in which we uniformly control DBC for each task. The goal of the proposed approach is to estimate a good meta-parameter such that the summation of empirical risks for each task is minimized and meanwhile each task is fair.

\subsection{Settings}
In this work, we consider a collection of supervised learning tasks $\mathcal{T} = \{(\mathcal{D}_j^{S}, \mathcal{D}_j^{Q})\}_{j=1}^T$ which distributions over $\mathcal{Z}$ and $T$ is denoted as the number of tasks. $\mathcal{T}$ is often referred to as a meta-training set as well as an episode $(\mathcal{D}_j^{S}, \mathcal{D}_j^{Q})$ explicitly contains a pair of a support $\mathcal{D}_j^{S}$ and a query $\mathcal{D}_j^{Q}$ data sets. For each task $j\in\{1,2,...,T\}$, we let $\{\mathbf{x}_{j,i}, y_{j,i}, s_{j,i}\}_{i=1}^m \in(\mathcal{X\times Y\times S})$ be the corresponding task data and $m$ is the number of datapoints in the support set. For example, standard few-shot learning benchmarks evaluate model in $N$-way $K$-shot classification tasks and thus $m=N\times K$ indicates, in the support set of the $j$-th task, it contains $N$ categories and each consists of $K$ datapoints. We emphasize that we need to sample without replacement, \textit{i.e.,} $\mathcal{D}_j^S \cap\mathcal{D}_j^Q=\emptyset$.

In a general meta-learning setting, it consists of meta-train and meta-test partitions where each contains a number of mini-batches of episodes (see Figure \ref{fig:overvoew}). We consider a distribution over tasks $p(\mathcal{T})$ that we want our model to be able to adapt to. In a $N$-way-$K$-shot learning setting, a task $\mathcal{T}_j$ is sampled from $p(\mathcal{T})$, where the subscript $j$ represents the $j$-th task of a mini-batch. In the supervised learning setting, supposing the meta-model is a parameterized function $f_\phi$ with parameters $\phi$. In a general meta-learning model, the goal is to learn an optimized $\phi$ so that the summation of query losses $l_{\mathcal{T}_j}(f_\phi)$ over all meta-training tasks is minimum. During meta-training, $\phi$ is updated iteratively.

\begin{align}
    \phi^* = \arg\min_\phi \mathbb{E}_{\mathcal{T}\sim p(\mathcal{T})} l_{\mathcal{T}}(f_\phi)
\end{align}

\subsection{Model-Agnostic Meta-Learning with convex constraints}
Meta-learning approaches for few-shot learning are often assumed that the support and query sets of a task are sampled from the same distribution. In our work, for each single task, the objective is to minimize the predictive error $\mathcal{L}^{inner}(\mathcal{D}^S_j,\phi)$ such that it is constrained by a function $g_j(\phi)$.

\begin{align}
\label{inner_problem}
    \min_{\phi_j} \quad &\mathcal{L}^{inner}(\mathcal{D}^S_j,\phi)  \\
    \text{subject to}\quad &g_j(\phi) \leq 0 \nonumber
\end{align}

where $\mathcal{L}^{inner}: \mathbb{R}^n \rightarrow \mathbb{R}$ is a loss function, such as cross-entropy for classification, and $g: \mathbb{R}^n \rightarrow \mathbb{R}$ is an appropriate complexity function ensuring the existence and the uniqueness of the above minimizer. A point $\phi$ in the domain of the problem is feasible if it satisfies the constraint $g_j(\phi)\leq 0$. Specifically, $g_j(\phi)$ is defined by the definition of decision boundary covariance in Eq.(\ref{dbc definition}), \textit{i.e.}
\begin{align}
    g_j(\phi) = \abs*{\frac{1}{N\times K}\sum_{\mathbf{s}_i, \mathbf{x}_i\sim\mathcal{T}_j}(\mathbf{s}_i-\Bar{\mathbf{s}})d_\alpha(\mathbf{x}_i)}-c
\end{align}
where $c$ is a small positive fairness relaxation, $d_\alpha(\mathbf{x}_i)\approx\max\{\mathbf{p}_n\in[0,1]^N\}$ and $\mathbf{p}$ denotes the class probabilities of $\mathbf{x}_i$. Here, for a $N$-way-$K$-shot classification task, we include $N\times K$ data points in the support set $\mathcal{D^S}$ of each task $\mathcal{T}_j$. 

To solve the optimization problem, we thus introduce an unified Lagrange multiplier $\lambda\geq 0$ for all tasks and the Lagrange function $\mathcal{L}_{\mathcal{T}_j}(\phi, \lambda)$  for each task is defined by 

\begin{align}
    \mathcal{L}_{\mathcal{T}_j}(\phi, \lambda) = \mathcal{L}^{inner}(\phi) + \lambda(g_j(\phi))
\end{align}

Therefore the original problem can be finally seen by minimizing $\mathcal{L}_{\mathcal{T}_j}(\phi, \lambda)$ for each task and thus mitigates dependency of prediction on the protected attribute. 
The goal of training a single task is to output a local parameter $\phi_j$ given the meta-parameter $\phi$ such that it minimizes the task loss $\mathcal{L}^{inner}$ subject to the task constraint $g_j(\phi)\leq 0$. Next, to update the meta-parameter, we minimize the generalization error $\mathcal{L}^{meta}$ using query sets across every task in the batch such that the query constraints are satisfied. Formally, the learning objective across all tasks is

\begin{align}
\label{meta_problem}
    \min_{\phi} \quad &\mathcal{L}^{meta}(\sum_{j=1}^T\mathcal{D}_j^Q,\phi) =\sum_{j=1}^T \mathcal{L}^{inner}(\mathcal{D}^Q_j,\phi_j)
\end{align}

where $\phi_j=\arg\min_{\phi_j, g_j(\phi)\leq0}\mathcal{L}^{inner}$ is the local optimum for each task. A step-by-step learning algorithm for unfairness prevention in few-shot regression is proposed in Algorithm \ref{alg:bias prevention few-shot}.

\begin{algorithm}[h]
\caption{Unfairness Prevention in Few-Shot Classification.}
\textbf{Require: } $p(\mathcal{T})$: distribution over tasks.\\
\textbf{Require: } $\alpha, \beta$: step size hyperparameters.\\
\textbf{Require: } $q$: inner gradient update steps.
\begin{algorithmic}[1]
\State Randomly initialize $\phi$
\While{not done}
    \State Sample batch of tasks $\mathcal{T}_j$
    \For{all $\mathcal{T}_j = \{\mathcal{D}_j^\mathcal{S}, \mathcal{D}_j^\mathcal{Q}\}$}
        \State Sample $N$-way-$K$-shot datapoints from $\mathcal{D}_j^\mathcal{S}=\{\mathbf{x}^j, y^j, s^j\}$
        \State $\phi_j \leftarrow \phi$
        \For{$q=1,2,...$}
            \State Evaluate $\nabla_{\phi_j} \mathcal{L}_{\mathcal{T}_j}(\phi_j, \lambda)$ using $\mathcal{D}_j^\mathcal{S}$
            \State Compute adapted local parameter $\phi_j \leftarrow \phi_j-\alpha \nabla_{\phi_j} \mathcal{L}_{\mathcal{T}_j}(\phi_j, \lambda)$ 
        \EndFor
        \State Sample datapoints from $\mathcal{D}_j^\mathcal{Q}=\{\mathbf{x}^j, y^j, s^j\}$
        \State Evaluate query loss $l_{\mathcal{T}_j}(\phi_j)$ and query fairness $g_{\mathcal{T}_j}(\phi_j)$ using $\mathcal{D}_j^\mathcal{Q}$
    \EndFor
    \State Update $\phi \leftarrow \phi - \beta\nabla_\phi\sum_{\mathcal{T}_j\sim p(\mathcal{T})} l_{\mathcal{T}_j}(\phi_j)$
    \State Evaluate training fairness $mean(g_{\mathcal{T}_j}(\phi_j))$
\EndWhile
\end{algorithmic}
\label{alg:bias prevention few-shot}
\end{algorithm}

\subsection{Algorithm Analysis}
Since the proposed Algorithm \ref{alg:bias prevention few-shot} is modified following \cite{Finn-ICML-2017-(MAML)}, the convergence is guaranteed and detailed analysis is stated in \cite{Fallah-AISTATS-2020}. Accessing to sufficient samples, the running time of the algorithm is $O(n\cdot b\cdot q)$ , where $n$ is the number of outer iterations, $b$ is batch-size, and $q$ is gradient steps of inner loop. For a $N$-way-$K$-shot learning, the best accuracy is achieved when $||\nabla \phi||\leq O(\Tilde{\sigma}/\sqrt{NK})$, where $\phi = \mathbb{E}_{\mathcal{T}\sim p(\mathcal{T})} l_{\mathcal{T}}(f_\phi)$, $\sigma$ is a bound on the standard deviation of $\nabla \mathcal{L}_{\mathcal{T}_j}(\phi_j, \lambda)$ from its mean $\nabla \mathcal{L}_{\mathcal{T}}(\phi, \lambda)$, and $\Tilde{\sigma}$ is a bound on the standard deviation of estimating $\nabla \mathcal{L}_{\mathcal{T}_j}(\phi_j, \lambda)$ using a single data point.

\section{Experiments}
To validate our approach of unfairness prevention in few-shot meta-learning models, we conduct experiments 
with two real-world image data sets.

\subsection{Data}
\textbf{Omniglot} \cite{Lake-CogSci-2011} is a data set of 1623 handwritten characters collected from 50 alphabets. To avoid overfitting, data augmentation is used on images in the form of rotations of 90 degrees increments, \textit{i.e.} 90, 180, and 270 degrees. Rotated class samples are considered new classes and thus we have 1623$\times$4 classes in total. We shuffle all character classes and randomly split the data set into three sets, 1150$\times$4 for the training set, 50$\times$4 for validation, and 423$\times$4 for testing. We follow the procedure of \cite{Vinyals-NIPS-2016-(MatchingNet)} by resizing the grayscale images to 28$\times$28. In order to study fairness, an arbitrary probability $p(\mathbf{x}|s=1)$ is assigned to each class and thus each image sample is given a protected attribute $s\in\{0,1\}$. 

\begin{table}[!htbp]
\normalsize
    \caption{Key Characteristics of Experimental Data}
    \centering
    \begin{tabular}{p{3cm}|c|c|c}
        \hline
         & \multicolumn{2}{|c|}{\textbf{Omniglot}} & \textbf{mini-ImageNet}\\
        \hline
        images & \multicolumn{2}{|c|}{grey} & color \\
        \hline
        augmentation & \multicolumn{2}{|c|}{yes} & no \\
        \hline
        scale & \multicolumn{2}{|c|}{\textbf{$28\times 28$}} & $84\times 84$\\
        \hline
        classes for training & \multicolumn{2}{|c|}{\textbf{$1150\times 4$}} & $64$\\
        \hline
        classes for validation & \multicolumn{2}{|c|}{\textbf{$50\times 4$}} & $12$\\
        \hline
        classes for test & \multicolumn{2}{|c|}{\textbf{$423\times 4$}} & $24$\\
        \hline
        $N$-way & 5 & 20 & 5\\
        \hline
        local gradient step(s) for training & 1 & 5 & 5\\
        \hline
        local gradient step(s) for evaluation & 3 & 5 & 10\\
        \hline
        local learning rate for training & 0.4 & 0.1 & 0.01\\
        \hline
        local learning rate for evaluation & 0.4 & 0.1 & 0.01\\
        \hline
        meta batch-size & 32 & 16 & 4 (1-shot), 2 (5-shot)\\
        \hline

    \end{tabular}
    \label{tab:dataset overview}
\end{table}

\textbf{mini-ImageNet} is originally proposed by \cite{Vinyals-NIPS-2016-(MatchingNet)}. It consists of 60,000 color images scaled down to 84$\times$84 divided into 100 classes with 600 examples each. We use the split proposed in \cite{Ravi-ICLR-2017}, which consists of 64 classes for training, 12 classes for validation and 24 classes for testing. The protected attribute is randomly added following the same procedure stated in our Omniglot settings.

\subsection{Parameter Tuning}

In order to provide a fair comparison for all methods, our embedding architecture mirrors that used by \cite{Vinyals-NIPS-2016-(MatchingNet)}. It consists of four modules and each comprises a 64-filter $3\times 3$ convolution, batch normalization layer \cite{Ioffe-arXiv-2015}, a ReLU nonlinearity and a $2\times 2$ max-pooling layer. For Omniglot, since each image is resized to $28 \times 28$, it results in a 64-dimensional output space. Due to the increased size of images in mini-ImageNet, the resulting feature map is 1600-dimensional. All models are trained with Adam optimizer. 

For $N$-way, $K$-shot classification, each gradient is computed using a batch size of $N\times K$ examples. For Omniglot, the 5-way convolutional model is trained with 1 gradient step with step size of $\alpha=0.4$, and a meta batch-size of 32 tasks. The network is evaluated using 3 gradient steps with the same step size $\alpha=0.4$. The 20-way convolutional model is trained and evaluated with 5 gradient steps with step
size of $\alpha=0.1$. During training, the meta batch-size is set to 16 tasks. For MiniImagenet, the model is trained using 5 gradient steps of size $\alpha=0.01$, and evaluated using 10 gradient steps at test time. Following \cite{Ravi-ICLR-2017}, 15 examples per class are used for evaluating the post-update meta-gradient. We used a meta batch-size of 4 and 2 tasks for 1-shot and 5-shot training respectively. All model are trained for 60000 iterations.

Besides, to comply with disparate treatment criterion, we specify that protected attributes are not used in the decision making, \textit{i.e.} protected and explanatory attributes consist of disjoint feature sets. Key characteristics for all data set are listed in Table \ref{tab:dataset overview}.

\subsection{Baseline Methods}
We compared our work with three well known meta-learning state-of-the-arts, MAML \cite{Finn-ICML-2017-(MAML)}, Matching Networks \cite{Vinyals-NIPS-2016-(MatchingNet)}, and Prototypical Networks \cite{Snell-NIPS-2017-(ProtoNet)}. These methods apply the same meta-learning framework but differ in their strategies to make predictions conditioned on the support set. MAML is a gradient based meta-learning algorithm, where each support set is used to adapt a uniformed initialized model parameters using one or few gradient steps. After several updates, the meta-loss reaches the minimum along with each episode loss reaching its local minimum. 

For both Matching and Prototypical Networks (\textit{i.e.} MatchingNet and ProtoNet), the prediction of the samples in a query set is based on comparing the distance between embedded query feature and support feature within each class. MatchingNet was the first to both train and test on $N$-way-$K$-shot tasks. The appeal of this is training and evaluating on the same tasks lets us optimise for the target task in an end-to-end fashion. Different from earlier approaches such as siamese networks \cite{Koch-ICML-2015}, MatchingNet combines both embedding and classification to form an end-to-end differentiable nearest neighbours classifier. Specifically, it applied a bidirectional Long-Short Term Memory (LSTM) to encode in the context of the support set and compares cosine distance between the query feature and each support feature. 

In Prototypical Networks, the authors apply a compelling inductive bias in the form of class prototypes to achieve impressive few-shot performance that exceeds Matching Networks without the complication of full context embeddings or FCE for short. The key assumption is that there exists an embedding in which samples from each class cluster around a single prototypical representation which is simply the mean of the individual samples. This idea streamlines $K$-shot classification in the case of $K > 1$ as classification is simply performed by taking the label of the closest class prototype.

In order to output fair predictions, fairness constraints are applied. Experimental results shown with prefix ``Fair-" in the front indicate models adjusted using our proposed approach.

\begin{table*}[!htbp]
\normalsize
    \centering
    \captionof{table}{Consolidated overall result for few-shot classification accuracies and fairness. Methods with superscript $^*$ and $\ddagger$ respectively indicates results reported by the original paper and our local replication with addition of protected attributes.}
    \begin{tabular}{l|cc|cc|cc|cc}
        \hline
        \multirow{2}{*}{\textbf{Omniglot}} & \multicolumn{4}{c|}{\textbf{5-way}} & \multicolumn{4}{c}{\textbf{20-way}}\\
        & \multicolumn{2}{c}{\textbf{1-shot}} & \multicolumn{2}{c|}{\textbf{5-shot}} & \multicolumn{2}{c}{\textbf{1-shot}} & \multicolumn{2}{c}{\textbf{5-shot}}\\
        \hline
        \textbf{Approach} & \textbf{Accuracy} & \textbf{DBC} & \textbf{Accuracy} & \textbf{DBC} & 
        \textbf{Accuracy} & \textbf{DBC} &
        \textbf{Accuracy} & \textbf{DBC}\\
        \hline
        MAML$^*$ & 98.7$\pm$0.4\% & - & 99.9$\pm$0.1\% & - & 95.8$\pm$0.3\% & - & 98.9$\pm$0.2\% & - \\
        MAML$^\ddagger$ & 98.1$\pm$0.5\% & 0.39$\pm$0.01 & 98.1$\pm$0.1\% & 0.59$\pm$0.09 & 96.2$\pm$0.6\% & 0.50 $\pm$0.10 & 98.7$\pm$0.4\% & 0.59$\pm$0.03\\
        Fair-MAML & 89.4$\pm$0.5\%  & 0.18$\pm$0.01 & 92.2$\pm$0.4\% & 0.19$\pm$0.01 & 85.4$\pm$0.3\%  & 0.21$\pm$0.03 & 90.4$\pm$0.4\% & 0.27 $\pm$0.01\\
        \hline
        MatchingNet$^*$ & 98.1\% & - & 98.9\% & - & 93.8\% & - & 98.7\% & -\\
        MatchingNet$^\ddagger$ & 98.0$\pm$0.4\% & 0.45$\pm$0.02 & 98.6$\pm$0.3\% & 0.45$\pm$0.01 & 95.2$\pm$0.5\% & 0.47$\pm$0.02 & 97.9$\pm$0.3\% & 0.48$\pm$0.03 \\
        Fair-MatchingNet & 96.5$\pm$0.5\% & 0.14$\pm$0.01 & 98.0$\pm$0.5\% & 0.15$\pm$0.01 & 94.1$\pm$0.5\% & 0.13$\pm$0.03 & 98.5$\pm$0.5\% & 0.13$\pm$0.01 \\
        \hline
        ProtoNet$^*$ & 98.8\% & - & 99.7 \% & - & 96.0\% & - & 98.9\% & - \\
        ProtoNet$^\ddagger$ & 98.5$\pm$1.8\% & 0.44$\pm$0.01  & 99.1$\pm$0.3\% & 0.44$\pm$0.03 & 96.6$\pm$1.7\% & 0.46$\pm$0.01 & 98.6$\pm$0.1\% & 0.48$\pm$0.01\\
        Fair-ProtoNet & 86.4$\pm$2.0\% & 0.10$\pm$0.07 & 94.3$\pm$0.5\% & 0.10$\pm$0.02 & 88.7$\pm$1.5\% & 0.03$\pm$0.02& 89.4$\pm$0.5\% & 0.02$\pm$0.01\\
        \hline
    \end{tabular}
    \label{tab:Omniglot}
\end{table*}
\begin{table*}[!htbp]
\normalsize
    \centering
    \begin{tabular}{l|cc|cc}
        \hline
        \textbf{mini-ImageNet}\quad \textbf{5-way} & \multicolumn{2}{c}{\textbf{1-shot}} & \multicolumn{2}{c}{\textbf{5-shot}}\\
        \hline
        \textbf{Approach} & \textbf{Accuracy} & \textbf{DBC} & \textbf{Accuracy} & \textbf{DBC} \\
        \hline
        MAML$^*$ & 48.7$\pm$1.8\% & - & 63.1$\pm$0.9\% & -\\
        MAML$^\ddagger$ & 44.2$\pm$1.1\% &0.91$\pm$0.01 & 61.1$\pm$0.8\% & 0.89$\pm$0.01 \\
        Fair-MAML &35.5$\pm$0.9\% &0.61$\pm$0.05 &54.5$\pm$0.8\% & 0.58$\pm$0.07 \\
        \hline
        MatchingNet$^*$ & 43.6$\pm$0.8\% & - & 55.3$\pm$0.7\% & - \\
        MatchingNet$^\ddagger$ & 44.8$\pm$0.1\% & 0.10$\pm$0.01 & 60.9$\pm$1.0\% & 0.13$\pm$0.03 \\
        Fair-MatchingNet  & 37.1$\pm$2.0\% & 0.06$\pm$0.05 & 56.5$\pm$2.0\% & 0.07$\pm$0.05\\
        \hline
        ProtoNet$^*$ & 49.4$\pm$0.8\% & - & 68.2$\pm$0.7\% & - \\
        ProtoNet$^\ddagger$ & 43.3$\pm$2.9\% & 0.09$\pm$0.03 & 64.2$\pm$1.0\% & 0.14$\pm$0.01\\
        Fair-ProtoNet & 39.9$\pm$1.0\% & 0.05$\pm$0.02 & 59.9$\pm$2.0\% & 0.06$\pm$0.05 \\
        \hline
    \end{tabular}
    \label{tab:mini-ImageNet}
\end{table*}

\subsection{Experimental Results}
Table \ref{tab:Omniglot} showcases results experimented through three cutting-edge meta-learning methods and those associated with our proposed unfairness prevention approach (noted with ``Fair-"), which examined with two real-world image data sets, \textit{i.e.} Omniglot \cite{Lake-CogSci-2011} and mini-ImageNet \cite{Vinyals-NIPS-2016-(MatchingNet)}. The problem of $N$-way classification is set up as follows: select $N$ unseen classes, provide the model with $K$ different instances of each of the $N$ classes, and evaluate the model’s ability to classify new instances within the $N$ classes.

In order to check the generalized fairness of these state-of-the-arts on unseen tasks, we produce local replications labeled with superscript $\ddagger$ that are used to compare with the results reported in the original paper labeled with $^*$. Note that for local replication methods, input images are slightly different, where in this paper we additionally consider the protected attribute as one of the input features. Besides, in the proposed unfairness prevention approach (labeled with the prefix ``Fair-" in Table \ref{tab:Omniglot}), cross-entropy losses are calculated with using images without the protected attribute, as the fairness constraint is applied to control the covariance between the protected variable and the signed distance from the feature vectors to the decision boundary.

Besides, our approaches (Fair-MAML, Fair-MatchingNet, and Fair-ProtoNet) are outperformed than the original methods and the gap between all methods is narrowing as the number of training data increases (\textit{i.e.} from $1$-shot to $5$-shot). However, in Omniglot data, as more image classes are considered in the experiment, accuracy decreases. Even in the 20-way classification problem, our approach is able to control the decision boundary covariance efficiently.

Our proposed approach is empirically shown to reduce DBC and thus improve outcome fairness in multi-class decision-making for all selected meta-learning methods. Another notable observation is that decreasing unfairness is brought at the sacrifice of classification accuracy in a bit. This comes down to the trade-off between accuracy and fairness. In summary, our approach significantly controls biases for few-shot meta-learning models and generalizes to unseen tasks.

\section{Conclusion and Future Work}
In this paper, for the first time we develop deep into the few-shot supervised meta-learning classification model and propose a bias-control approach by adding statistical parity constraint, namely decision boundary covariance, which significantly mitigates dependence of prediction on the protected variable in each task and generalize both accuracy and fairness to unseen tasks. Experimental results on two real-world image data sets indicate the proposed approach works for three cutting-edge few-shot meta-learning models with multi-class classification problems. Further research in this area can make multitask parameters a standard ingredient in deep fairness learning.

\section*{Acknowledgement}
This work is supported by the National Science Foundation (NSF) under Grant No \#1815696 and \#1750911.

\bibliographystyle{IEEEtran}
\bibliography{references}

\end{document}